\documentclass{article}

\PassOptionsToPackage{numbers,sort&compress}{natbib}
\usepackage[preprint]{neurips_2026}

\usepackage[utf8]{inputenc}
\usepackage[T1]{fontenc}
\usepackage{hyperref}
\hypersetup{colorlinks=true, citecolor=blue, linkcolor=blue, urlcolor=blue}
\usepackage{url}
\usepackage{booktabs}
\usepackage{amsmath}
\usepackage{amsfonts}
\usepackage{microtype}
\usepackage{xcolor}
\usepackage{graphicx}

\makeatletter
\renewcommand{\@notice}{}  % no "Preprint." line at the foot of the first page
\renewcommand{\paragraph}{%
  \@startsection{paragraph}{4}{\z@}%
                {1.5ex \@plus 0.5ex \@minus 0.2ex}%
                {-0.5em}%
                {\normalsize\bf}%
}
\makeatother

\newcommand{\ours}{G0.7}
\newcommand{\oursk}{G0.7-20\%}
\newcommand{\oursr}{G0.7-Robot}
\newcommand{\pifive}{$\pi_{0.5}$}

\title{G0.7: Scaling Up a Fast World Action Model\\with Long Context}

\author{Galaxea Team\\
  \url{https://www.galaxea-ai.com/model-g07.html}}

\begin{document}

\maketitle
\suppressfloats[t]  % nothing floats above the title

\begin{abstract}
World action models (WAMs) adapt pretrained video generators for robot control by jointly predicting future camera frames and actions, transferring physical priors learned from video into manipulation policies. However, generating future frames before every action makes test-time inference too slow for closed-loop control, while most existing policies remain restricted to short observation histories. Building on Fast-WAM, which showed that video prediction shapes policy representations during training so that future frame synthesis can be skipped at deployment, we present \ours{}, a generalist world action model scaled along two axes: extending the context window from a single observation to up to five minutes of history, and scaling pretraining to a large-scale corpus of robot teleoperation and human egocentric (ego) video. Across public simulation suites and real-robot deployments, direct action decoding (Fast mode) matches imagine-then-act (IDM) inference from the same checkpoint while running in 60\,ms per 16-step action chunk on a single RTX 5090 (7.6$\times$ faster than IDM) and supporting pause-free real-time chunking. Performance scales consistently with pretraining data, with human ego video driving the largest gains in out-of-distribution generalization: \ours{} achieves a 70.8 mean score across five simulation benchmark suites, 48\% and 54\% zero-shot pick-and-place success on Galaxea R1 Pro and R1 Lite, 96.7\% in-distribution and 93.3\% out-of-distribution success on video-conditioned in-context learning tasks, and 86.9\% mean task progress across seven real-world post-training tasks including multi-minute box folding, drink preparation, and product bagging.
\end{abstract}

\section{Introduction}
\label{sec:intro}

Generalist robot manipulation requires policies that combine physical commonsense, low-latency closed-loop reactivity, and temporal memory across extended horizons. While vision-language-action models (VLAs)~\citep{pi0,pi05} initialize policies from pretrained vision-language models to inherit semantic reasoning, world action models (WAMs)~\citep{dreamzero,lingbot,vpp2,openwam,fastwam} build on pretrained video generators and jointly predict future camera frames alongside robot actions, transferring rich physical and spatiotemporal priors from video into control.

% Placed here, with [t], so that it lands at the top of page 2 instead of floating past the introduction to page 3.
\begin{figure}[t]
  \centering
  \includegraphics[width=\linewidth]{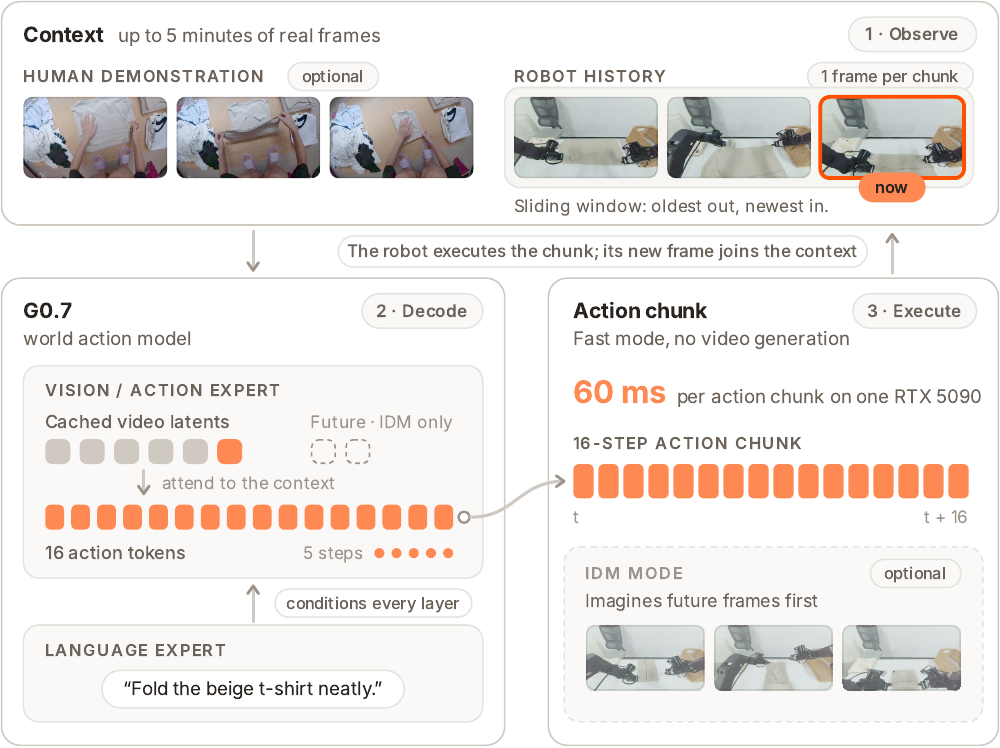}
  \caption{\ours{} architecture and control loop. The context caches up to five minutes of real camera observations, combining the robot's own history with an optional demonstration video. Conditioned at every layer by the language expert, the vision/action expert attends over this context to decode the next 16-step action chunk directly in 60\,ms (Fast mode), or optionally imagines future frames first (IDM mode). After each chunk executes, the newest frame joins the context and the oldest slides out of the window.}
  \label{fig:model}
\end{figure}

Despite their promise, existing WAMs face three practical limitations when deployed on physical robots. First, most WAMs synthesize future video frames at every control step before decoding actions~\citep{dreamzero,lingbot,vpp2}, incurring hundreds of milliseconds of latency per chunk and stalling closed-loop execution. 
Second, like most visuomotor policies, they typically condition on only one or a few recent frames, leaving the policy unable to track state over long horizons under partial observability or adapt to new tasks using human demonstrations as context. 
Third, publicly documented WAM pretraining corpora remain relatively modest in scale compared to the visual diversity required for zero-shot and out-of-distribution (OOD) generalization across tasks and embodiments.

In Fast-WAM~\citep{fastwam}, we addressed the first limitation by showing that test-time video generation is not what makes WAMs effective: video prediction performs its role during training by shaping the backbone's spatiotemporal representations, allowing the policy to skip future frame generation at test time and decode actions directly without losing control performance. However, Fast-WAM was a small-scale study restricted to single-frame observations and benchmark-specific datasets.

In this report, we introduce \ours{}, Galaxea's new generalist robot policy, which scales the Fast-WAM paradigm along both temporal context and pretraining data (Figure~\ref{fig:model}):
\begin{itemize}
  \item \textbf{Native long-context autoregressive modeling (Section~\ref{sec:long-context}).} \ours{} conditions on up to five minutes of real camera observations maintained as clean video latents in a sliding-window KV cache. Pretrained autoregressively over multi-chunk sequences and extended during downstream adaptation without auxiliary memory modules, \ours{} supports both long-horizon episode memory and video-based in-context learning (ICL), 
  where the model can follow a human demonstration provided as an in-context video prompt.
  \item \textbf{Fast path at scale (Section~\ref{sec:fast-mode}).} Through stochastic attention masking during training, each \ours{} checkpoint supports both imagine-then-act (IDM) and direct action decoding (Fast) from the same weights. At the full pretraining scale, Fast mode matches IDM across all four evaluation suites while running in 60\,ms per 16-step action chunk in FP8 on a single RTX 5090 (156\,ms vs.\ 457\,ms in unoptimized BF16 eager mode) and integrating directly with real-time chunking (RTC)~\citep{rtc} for continuous motion.
  \item \textbf{Large-scale robot and egocentric human pretraining (Section~\ref{sec:data}).} We pretrain \ours{} on a large-scale corpus of manipulation data with multi-embodiment robot teleoperation and first-person human egocentric videos. Our data pipeline pairs three-level hierarchical language annotations and motion quality filtering on robot trajectories with 3D hand-pose extraction on human video, mapping human wrists and fingertips into a unified end-effector action representation aligned with the robot gripper convention.
\end{itemize}
Extensive evaluations across five public simulation benchmark suites and three real-robot settings (Section~\ref{sec:experiments}) demonstrate that \ours{} outperforms leading VLA and WAM baselines, including \pifive{}~\citep{pi05}, OpenWAM~\citep{openwam}, G0.5~\citep{g05}, Dream-1~\citep{dream1}, and Long-WAM~\citep{longwam}, across simulation, zero-shot deployment, in-context learning, and real-world post-training on dexterous tool use and multi-minute production workflows.

\section{G0.7}
\label{sec:model}

\ours{} builds on a standard dual-expert world action model architecture (Figure~\ref{fig:model}). A language expert encodes the task instruction and visual context, conditioning at every layer a shared vision/action expert that operates over causal video VAE latents, proprioceptive states, and 16-step action chunks ($4{:}1$ action-to-video frequency ratio). Multi-embodiment robot trajectories and human hand trajectories are mapped into a unified state/action vector with per-embodiment dimension masking, and trained over chunk sequences under a standard joint video-action conditional flow-matching objective~\citep{fastwam} with chunk-causal attention over clean historical video latents. Below, we focus on the two design choices that distinguish \ours{} at scale: native long-context autoregressive modeling (Section~\ref{sec:long-context}) and dual-mode Fast/IDM training and inference (Section~\ref{sec:fast-mode}).

\subsection{Long-Context Autoregressive Modeling}
\label{sec:long-context}

While most existing VLAs and WAMs condition on only one or two recent frames, \ours{} conditions on up to five minutes of observation history, maintained as clean video latents in a sliding-window KV cache. Long-context conditioning is built into pretraining from the start: \ours{} is trained autoregressively over multi-chunk sequences with variable sequence lengths $K$ (with per-rank batch sizes balanced to equalize compute across short and long sequences), together with random positional chunk offsets so the rotary position embeddings generalize across extended rollouts. Because attention operates over a standard sliding KV cache of clean observation latents, the context window can be further extended during downstream adaptation without architectural changes.

\begin{figure}[!htb]
  \centering
  \includegraphics[width=\linewidth]{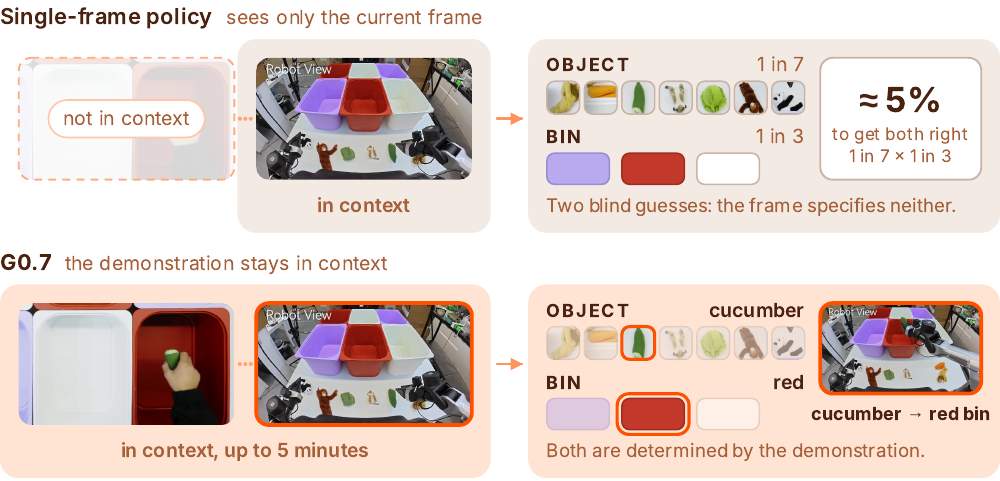}
  \caption{In-context task specification on sorting by example. From the current observation alone, the robot would have to guess both which of the seven objects to pick and which bin to put it in; attending to the human demonstration in context resolves both.}
  \label{fig:context}
\end{figure}

Beyond remembering earlier events within an episode, such as tracking an object after it is occluded, a five-minute context window enables in-context learning (ICL). Prepending a video of a person performing a task into the KV cache specifies the goal by example, allowing the robot to infer the underlying rule and continue the task autonomously. In sorting by example (Figure~\ref{fig:context}), the current camera view shows seven objects and three bins but gives no indication of which item to pick next or where it belongs; keeping the human demonstration in the KV cache resolves both decisions.

\subsection{Fast Mode: Action Decoding Without Video Generation}
\label{sec:fast-mode}

Every \ours{} checkpoint supports two inference paths from the same weights. During teacher-forced training, we apply per-sample stochastic attention masking ($p_{\text{IDM}} = 0.5$): action tokens either attend to the current chunk's future video latents (training the IDM path) or attend only to historical and current observations (training the Fast path), while video prediction remains supervised throughout. At inference time, IDM mode first denoises future video latents and then decodes actions conditioned on them, following the standard WAM paradigm; Fast mode skips video generation altogether and denoises the 16-step action chunk directly from the cached context in five steps.

\begin{figure}[!htb]
  \centering
  \includegraphics[width=\linewidth]{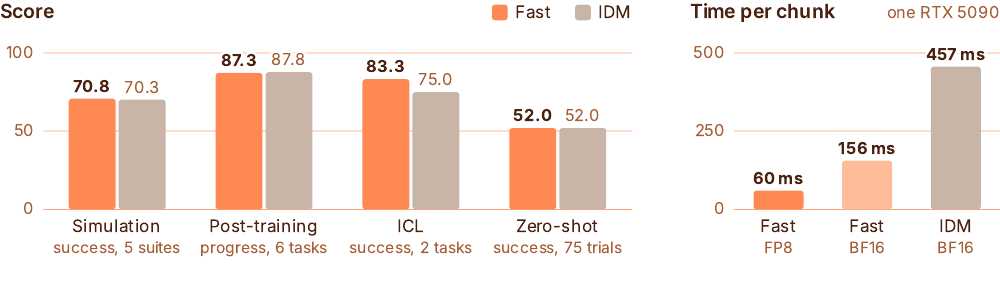}
  \caption{Fast mode vs.\ IDM mode from the same \ours{} checkpoint. Left: scores across the four evaluation suites of Section~\ref{sec:experiments}. Right: inference time per action chunk on one RTX 5090, comparing Fast mode in deployed FP8 against both modes in unoptimized BF16 eager mode.}
  \label{fig:fast-vs-idm}
\end{figure}

Evaluated on the same checkpoint, the two modes perform on par across all four evaluation suites of Section~\ref{sec:experiments} (Figure~\ref{fig:fast-vs-idm}). What changes is inference latency: on a single RTX 5090, Fast mode runs in 60\,ms per action chunk in our deployed FP8 configuration (7.6$\times$ faster than IDM), and in 156\,ms vs.\ 457\,ms (2.9$\times$ faster) when benchmarked head-to-head in unoptimized BF16 eager mode. Video prediction thus serves its purpose as a training objective, shaping the representations from which actions are decoded, while generating video again at test time adds latency without improving control.

Because Fast mode reduces inference to pure action denoising, it also makes \ours{} a drop-in fit for real-time chunking (RTC)~\citep{rtc}: the next action chunk is denoised while the current one executes, eliminating pauses between chunks.

\section{Pretraining Data}
\label{sec:data}

\begin{figure}[!htb]
  \centering
    \makebox[0.035\linewidth][c]{\rotatebox[origin=c]{90}{\scriptsize Robot}}%
  \raisebox{-.5\height}{\includegraphics[width=0.1565\linewidth]{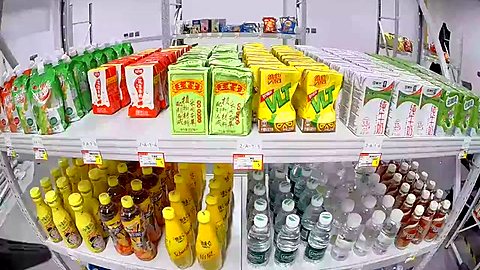}}\hfill
  \raisebox{-.5\height}{\includegraphics[width=0.1565\linewidth]{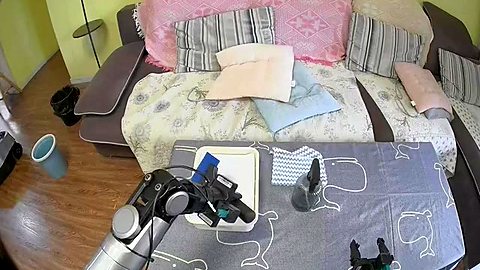}}\hfill
  \raisebox{-.5\height}{\includegraphics[width=0.1565\linewidth]{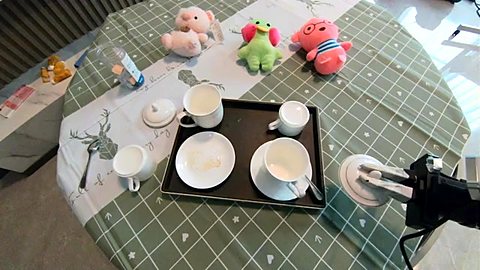}}\hfill
  \raisebox{-.5\height}{\includegraphics[width=0.1565\linewidth]{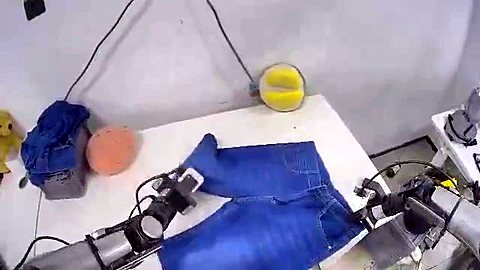}}\hfill
  \raisebox{-.5\height}{\includegraphics[width=0.1565\linewidth]{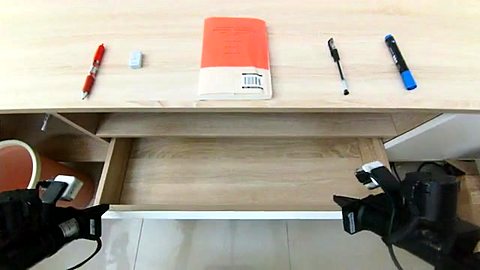}}\hfill
  \raisebox{-.5\height}{\includegraphics[width=0.1565\linewidth]{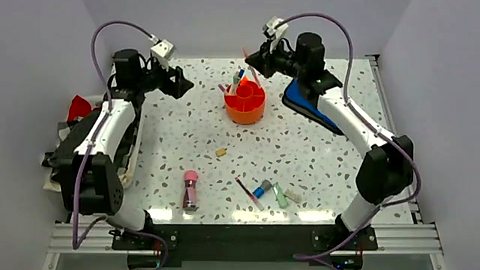}}\par\vspace{2pt}
    \makebox[0.035\linewidth][c]{\rotatebox[origin=c]{90}{\scriptsize Human ego}}%
  \raisebox{-.5\height}{\includegraphics[width=0.1565\linewidth]{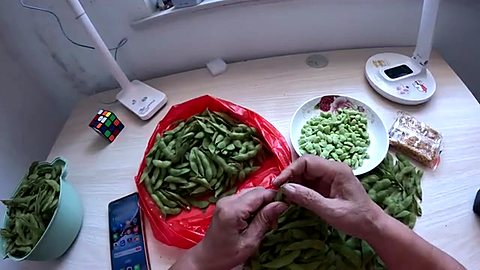}}\hfill
  \raisebox{-.5\height}{\includegraphics[width=0.1565\linewidth]{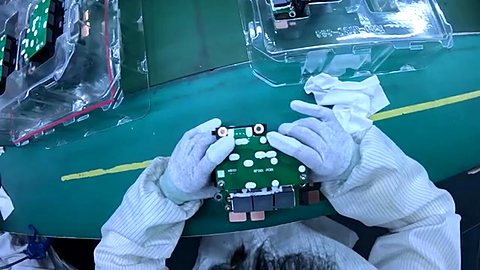}}\hfill
  \raisebox{-.5\height}{\includegraphics[width=0.1565\linewidth]{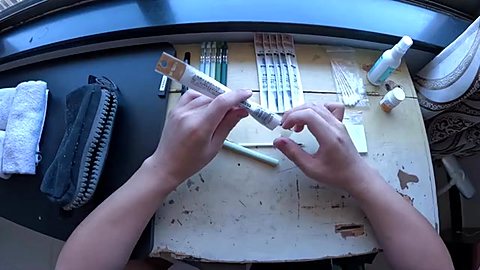}}\hfill
  \raisebox{-.5\height}{\includegraphics[width=0.1565\linewidth]{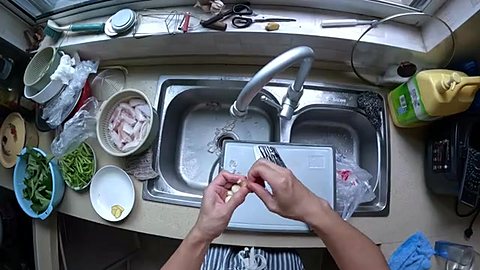}}\hfill
  \raisebox{-.5\height}{\includegraphics[width=0.1565\linewidth]{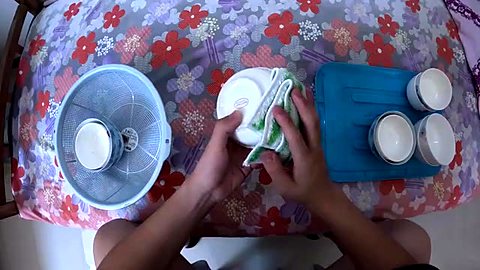}}\hfill
  \raisebox{-.5\height}{\includegraphics[width=0.1565\linewidth]{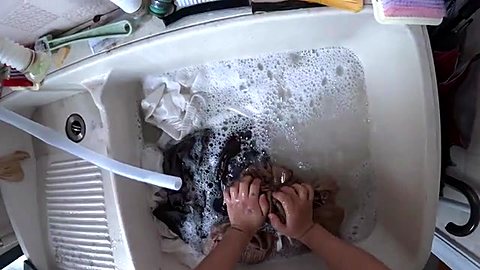}}\par\vspace{2pt}
  \caption{Sample frames from the pretraining data: robot teleoperation (top) and human ego video (bottom).}
  \label{fig:data}
\end{figure}

\ours{} is pretrained on a large-scale corpus of robot teleoperation and human ego video (Figure~\ref{fig:data}). The robot corpus combines trajectories from our Galaxea R1 Pro and R1 Lite fleets with public datasets including DROID~\citep{droid}, RoboMIND~\citep{robomind}, and BridgeData~\citep{bridgev2}. The ego corpus consists of first-person video of everyday manual work captured with head-mounted cameras.

\begin{figure}[!htb]
  \centering
  \includegraphics[width=\linewidth]{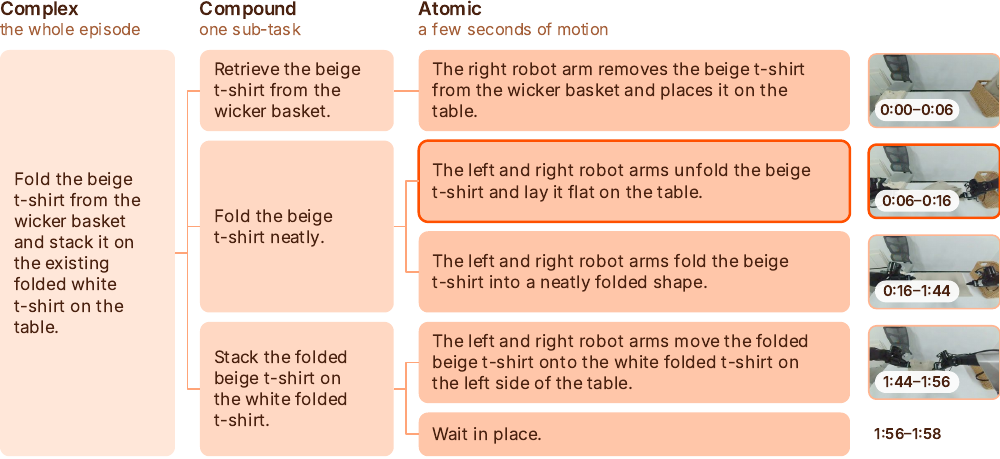}
  \caption{Three-level language annotation of a two-minute R1 Lite shirt-folding episode: one complex instruction for the full episode, three compound instructions for its sub-tasks, and five atomic labels, each paired with a representative frame and time span. Each training window is supervised at the matching temporal granularity. From 0:06 to 0:16 (outlined), the arms are unfolding the shirt, a step captured only by the atomic label.}
  \label{fig:annotation}
\end{figure}

\paragraph{Hierarchical language annotation.}
Teleoperation episodes are annotated at three temporal scales (Figure~\ref{fig:annotation}): a complex instruction for the full episode, compound instructions for sub-goals, and atomic labels covering a few seconds of motion. Relying on episode-level instructions alone creates a supervision mismatch. Early in a shirt-folding episode, for instance, the arms are unfolding the garment while the episode label says to fold and stack it. Supervising each training window at its matching granularity eliminates this mismatch between high-level language and low-level actions, and strengthens fine-grained instruction following so the same policy can execute both short atomic commands and multi-stage goals. Episodes also carry quality scores for motion smoothness and annotation reliability to filter out pause-heavy trajectories.

\begin{figure}[!htb]
  \centering
  \includegraphics[width=\linewidth]{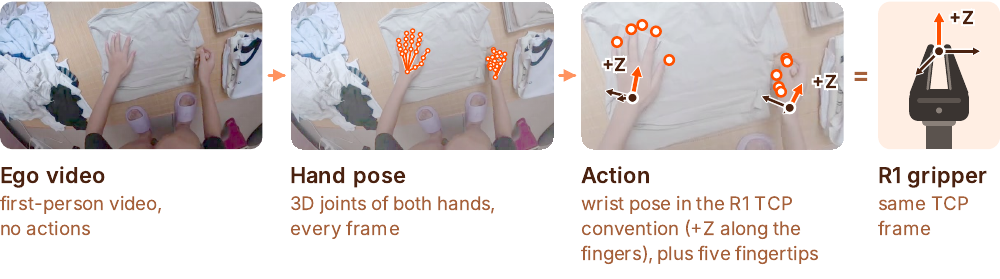}
  \caption{Extracting robot-compatible actions from human ego video. Estimated 3D hand joints in each frame are mapped to a wrist frame aligned with the R1 gripper TCP convention (+Z along the fingers) plus five fingertip positions.}
  \label{fig:ego-actions}
\end{figure}

\paragraph{Actions from ego video.}
To train on human ego video with the same objective as robot data, we built a custom data pipeline that extracts 3D hand trajectories from raw video and maps each hand into a unified end-effector action representation: the 6-DoF wrist pose aligned to the tool-center point (TCP) convention of the R1 gripper (+Z along the fingers) together with the 3D positions of all five fingertips (Figure~\ref{fig:ego-actions}). Human hands are treated as an additional embodiment under the joint video-action objective, while clips with noisy hand tracking still contribute to video prediction.

\section{Experiments}
\label{sec:experiments}

We evaluate \ours{} (running in Fast mode) across four settings: public simulation benchmarks, zero-shot deployment on real robots, in-context learning, and post-training on real-world manipulation tasks.

\paragraph{Baselines.}
Across these evaluations we compare against leading open-source and internal baselines: \pifive{}~\citep{pi05}, the most widely adopted open-source VLA; OpenWAM~\citep{openwam}, the state-of-the-art open-source WAM; and G0.5~\citep{g05}, our previous-generation VLA. In zero-shot experiments we also include Dream-1~\citep{dream1}, a recent WAM from Dream Labs designed for zero-shot bimanual control, and in in-context learning we compare against Long-WAM~\citep{longwam}, a WAM built specifically for long-horizon memory.

\paragraph{Checkpoints.}
To isolate the effect of pretraining scale and data composition, we compare three checkpoints that share the same architecture and differ only in their pretraining data (Table~\ref{tab:checkpoints}).

\begin{table}[!htb]
  \caption{Pretraining checkpoints compared in our data-scaling study.}
  \label{tab:checkpoints}
  \centering
  \begin{tabular}{ll}
    \toprule
    Checkpoint & Pretraining data \\
    \midrule
    \oursk{} & About 20\% of the robot teleoperation data \\
    \oursr{} & The full robot corpus, without human ego video \\
    \ours{}  & The full corpus, combining robot and human ego data \\
    \bottomrule
  \end{tabular}
\end{table}

\subsection{Simulation Benchmarks}
\label{sec:simulation}

We evaluate on five public benchmark suites that probe distinct capabilities. LIBERO~\citep{libero} tests tabletop manipulation alongside three robustness suites (LIBERO-Plus~\citep{liberoplus}, LIBERO-Pro~\citep{liberopro}, and LIBERO-Para~\citep{liberopara}) that perturb scene layout, task structure, and language instructions; we report the mean of these four as LIBERO w.\ OOD. RoboTwin 2.0~\citep{robotwin2} evaluates bimanual manipulation in the clean-to-random (C2R) setting under heavy domain randomization, VLABench~\citep{vlabench} focuses on semantic instruction following, RMBench~\citep{rmbench} tests memory-dependent manipulation, and RoboDojo~\citep{robodojo} spans 42 challenging tasks across generalization, memory, precision, and long-horizon control. Each model is post-trained on the respective benchmark's training split and evaluated under its standard protocol.

\begin{figure}[!htb]
  \centering
  \includegraphics[width=\linewidth]{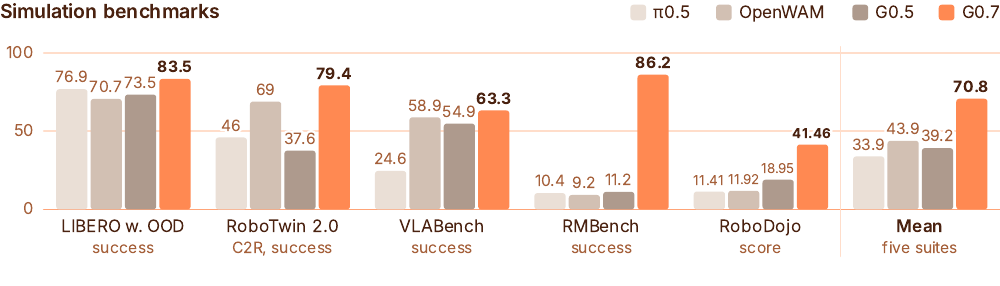}
  \caption{Simulation results after post-training on each benchmark. Scores are success rates in percent, except RoboDojo, which reports the benchmark's own score. LIBERO w.\ OOD is the mean of LIBERO, LIBERO-Plus, LIBERO-Pro, and LIBERO-Para.}
  \label{fig:simulation}
\end{figure}

Across the five benchmark suites, \ours{} achieves a mean score of 70.8, outperforming OpenWAM (43.9), G0.5 (39.2), and \pifive{} (33.9); Figure~\ref{fig:simulation} shows the per-suite results. The advantage is most pronounced on RMBench, where \ours{} reaches 86.2\% while single-frame baselines stay below 12\%. \ours{} also leads by a wide margin on RoboTwin 2.0 (79.4\% vs.\ 69.0\% for OpenWAM) and VLABench (63.3\% vs.\ 58.9\%). On LIBERO w.\ OOD, \ours{} reaches 83.5\% (vs.\ 76.9\% for \pifive{}): while standard LIBERO is near saturation across models (97\%--99\%), \ours{} separates itself on the perturbation suites, reaching 92.5\% on LIBERO-Plus (vs.\ 85.7\%) and 88.3\% on LIBERO-Para (vs.\ 71.4\%). On RoboDojo, \ours{} scores 41.46, more than double the strongest baseline (18.95 for G0.5).

\subsection{Zero-Shot Real-Robot Deployment}
\label{sec:zero-shot}

To test out-of-the-box generalization without any task-specific post-training, we deploy the pretrained checkpoints directly on real-world pick-and-place (PnP) benchmarks across two Galaxea platforms: 25 trials on R1 Pro and 50 trials on R1 Lite.

\begin{figure}[!htb]
  \centering
  \includegraphics[width=0.62\linewidth]{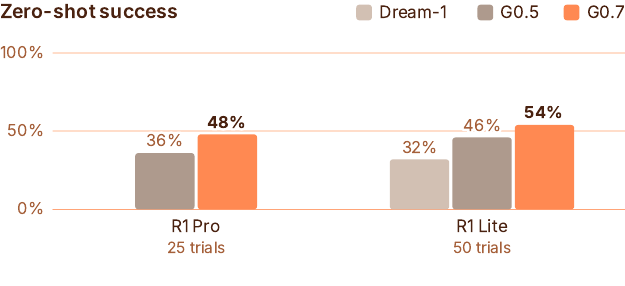}
  \caption{Zero-shot pick-and-place success rates on Galaxea R1 Pro (25 trials) and R1 Lite (50 trials). Dream-1 was evaluated on R1 Lite only.}
  \label{fig:zeroshot}
\end{figure}

\ours{} succeeds in 12 of 25 trials (48\%) on R1 Pro and 27 of 50 trials (54\%) on R1 Lite (Figure~\ref{fig:zeroshot}), outperforming both G0.5 (36\% and 46\%) and Dream-1 (32\% on R1 Lite). Zero-shot deployment is where human ego pretraining makes the sharpest difference: without ego data, \oursr{} succeeds in only 6 of 50 trials (12\%) on R1 Lite (Figure~\ref{fig:scaling}), confirming that visual and behavioral diversity from human video transfers directly to zero-shot robot control.

\subsection{In-Context Learning}
\label{sec:icl}

\begin{figure}[!htb]
  \centering
    \makebox[0.035\linewidth][c]{\rotatebox[origin=c]{90}{\scriptsize Shell game}}%
  \raisebox{-.5\height}{\includegraphics[width=0.2365\linewidth]{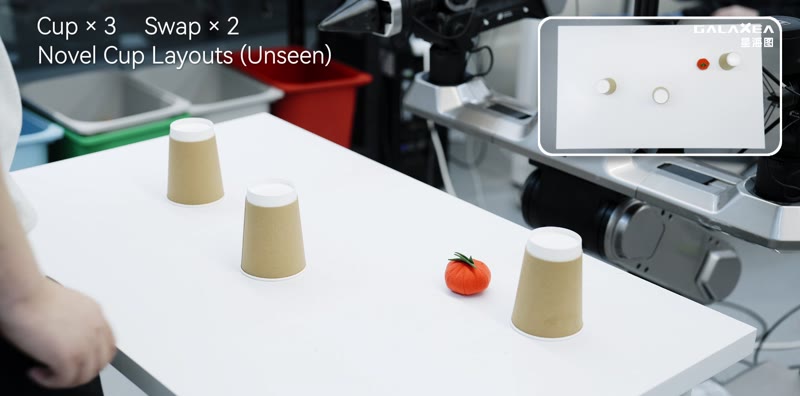}}\hfill
  \raisebox{-.5\height}{\includegraphics[width=0.2365\linewidth]{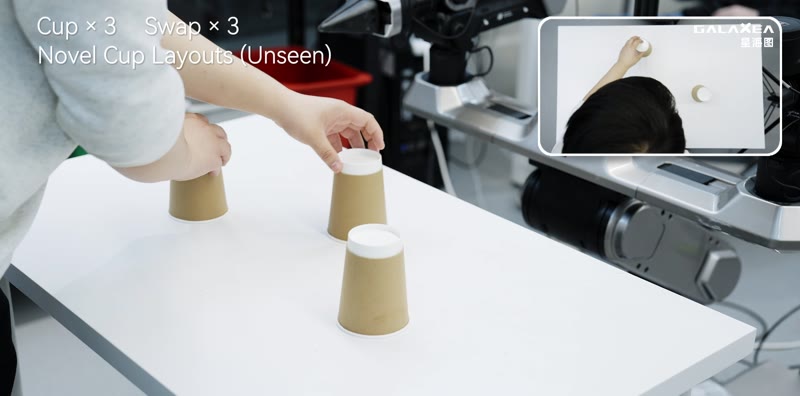}}\hfill
  \raisebox{-.5\height}{\includegraphics[width=0.2365\linewidth]{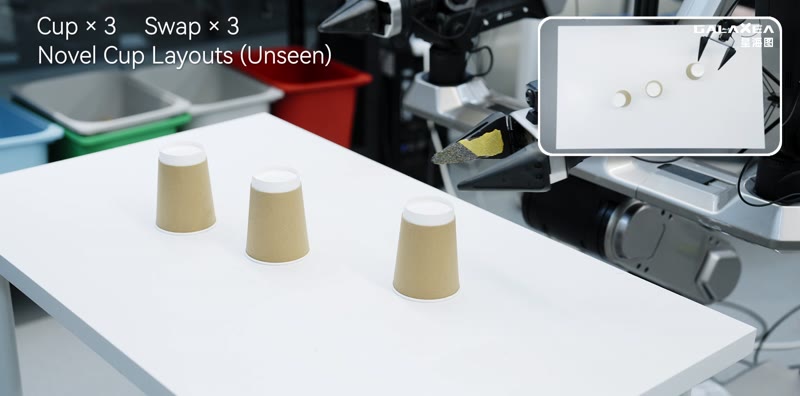}}\hfill
  \raisebox{-.5\height}{\includegraphics[width=0.2365\linewidth]{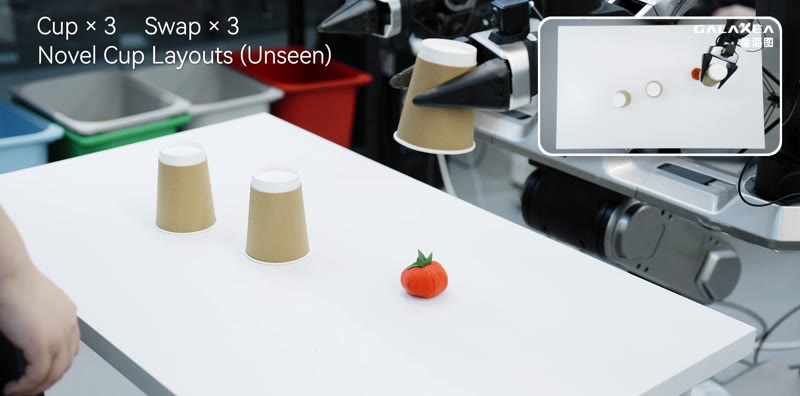}}\par\vspace{2pt}
    \makebox[0.035\linewidth][c]{\rotatebox[origin=c]{90}{\scriptsize Sorting}}%
  \raisebox{-.5\height}{\includegraphics[width=0.2365\linewidth]{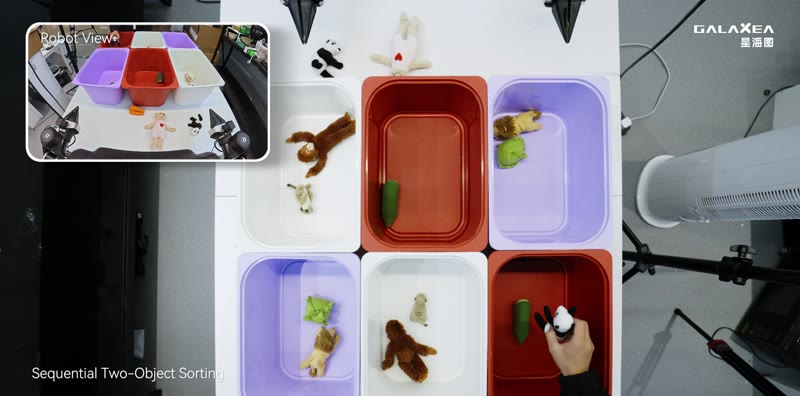}}\hfill
  \raisebox{-.5\height}{\includegraphics[width=0.2365\linewidth]{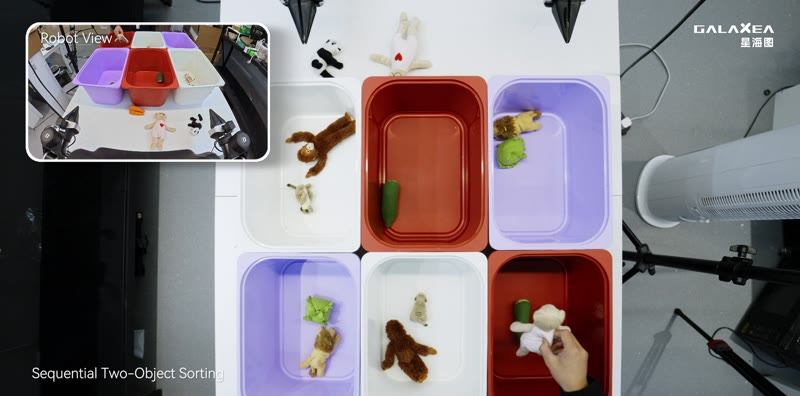}}\hfill
  \raisebox{-.5\height}{\includegraphics[width=0.2365\linewidth]{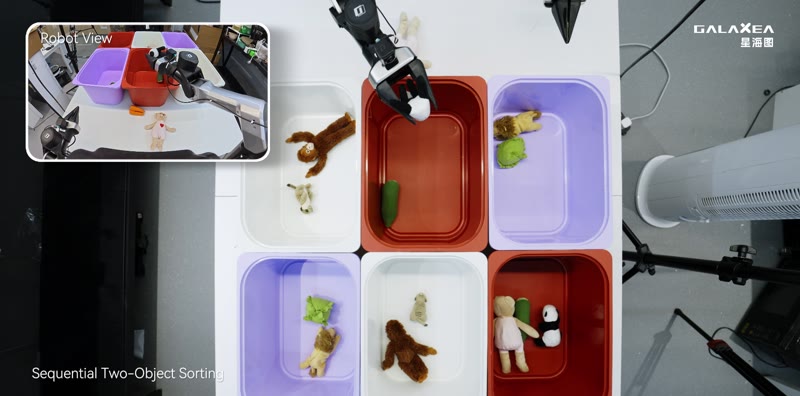}}\hfill
  \raisebox{-.5\height}{\includegraphics[width=0.2365\linewidth]{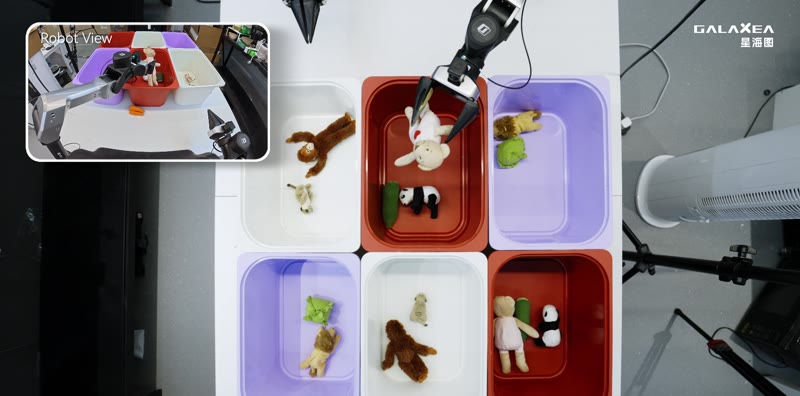}}\par\vspace{2pt}
  \caption{In-context learning tasks. \emph{Shell game} (top): a person hides an object under a cup and shuffles the cups; \ours{} tracks the hidden target and lifts the correct cup. \emph{Sorting by example} (bottom): a person shows where a few objects go, and \ours{} follows the human intent to sort objects into matching containers. Frames are from autonomous rollouts.}
  \label{fig:icl-rollouts}
\end{figure}

We next test whether \ours{} can act on information available only through extended visual contexts and generalize to unseen task configurations. We evaluate two complementary capabilities: \emph{long-horizon visual memory} and \emph{one-shot human demonstration following}. In the \emph{shell game}, the robot must track hidden targets through repeated shuffling. In \emph{sorting by example}, the robot must infer human intent from sparse cues in a single demonstration and generalize to unseen configurations (Figure~\ref{fig:icl-rollouts}).

\paragraph{Benchmark and generalization settings.}
Models are post-trained on task demonstrations and evaluated over 15 trials each in distribution (ID) and out of distribution (OOD). For the shell game, training and ID evaluation use three cups with 1--3 swaps, while OOD evaluation covers unseen cup types, target objects, layouts, cup counts (4--5 cups), shuffling styles, and longer sequences ($\geq$6 swaps). For sorting by example, training and ID evaluation focus on single-object sorting by demonstration, while OOD evaluation includes unseen objects, object counts, container colors and shapes, spatial layouts, and multi-object sequential sorting. Alongside standard post-training, we evaluate \textbf{ICL mid-training}, an intermediate stage between pretraining and task post-training that extends \ours{}'s context window and exposes the model to longer in-context demonstration sequences. We compare against Long-WAM, a competitive long-context WAM baseline, as well as G0.5 and \pifive{}, all post-trained under aligned settings with multi-frame visual contexts provided to G0.5 and \pifive{}.

\begin{figure}[!htb]
  \centering
  \includegraphics[width=\linewidth]{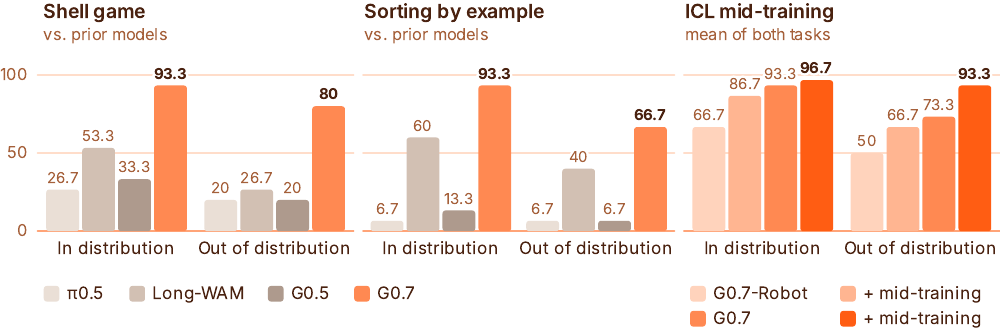}
  \caption{In-context learning success rates (\%) in distribution (ID) and out of distribution (OOD), evaluated over 15 trials per setting. Left and center: comparison with prior models on the shell game and sorting by example, respectively. Right: effect of ICL mid-training on \oursr{} and \ours{}, averaged across both tasks.}
  \label{fig:icl}
\end{figure}

\paragraph{Native long-context pretraining enables stronger downstream ICL.}
On the shell game, \ours{} achieves 93.3\% ID and 80.0\% OOD, outperforming Long-WAM (53.3\% / 26.7\%), G0.5 (33.3\% / 20.0\%), and \pifive{} (26.7\% / 20.0\%) (Figure~\ref{fig:icl}). On sorting by example, \ours{} reaches 93.3\% ID and 66.7\% OOD, compared with Long-WAM (60.0\% / 40.0\%), G0.5 (13.3\% / 6.7\%), and \pifive{} (6.7\% / 6.7\%). Despite receiving multi-frame visual contexts, G0.5 and \pifive{} remain substantially behind \ours{}, highlighting the importance of learning to effectively utilize extended visual histories rather than simply providing additional context frames.

\paragraph{Human-to-robot transfer via egocentric pretraining.}
Across both post-training settings, \ours{} substantially outperforms \oursr{} on OOD evaluation, improving mean OOD success from 50.0\% to 73.3\% without mid-training and from 66.7\% to 93.3\% (+26.7 percentage points) with ICL mid-training. These gains suggest that egocentric human video pretraining learns transferable representations of human interactions, facilitating the transfer of human intent and task knowledge to downstream robotic manipulation.

\paragraph{Effect of ICL mid-training.}
ICL mid-training further improves mean OOD success from 50.0\% to 66.7\% on \oursr{} and from 73.3\% to 93.3\% on \ours{}. Learning from paired human--robot demonstrations helps strengthen the correspondence between human intent and robot execution, enabling more reliable demonstration following across unseen configurations. Together with egocentric video pretraining, ICL mid-training brings \ours{} to 96.7\% ID and 93.3\% OOD success across both tasks.

\subsection{Real-World Post-Training}
\label{sec:post-training}

\begin{figure}[!htb]
  \centering
    \begin{minipage}[t]{0.494\linewidth}\centering
    \includegraphics[width=0.494\linewidth]{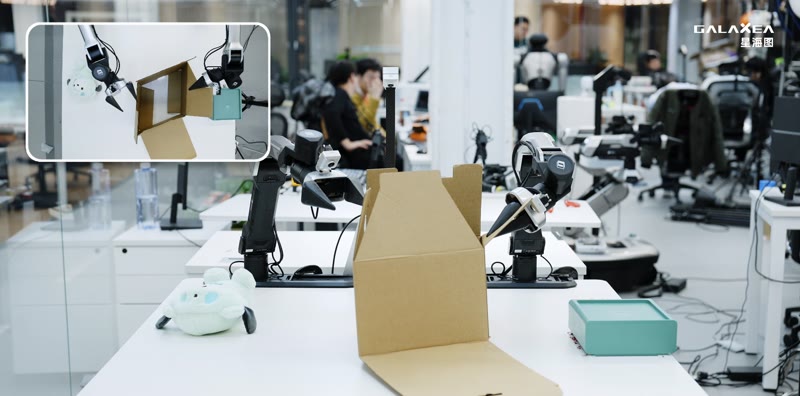}\hfill
    \includegraphics[width=0.494\linewidth]{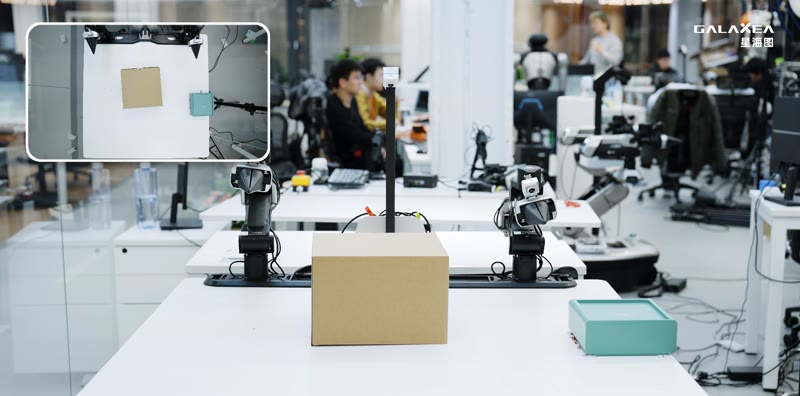}\\[1pt]
    {\footnotesize Fold a shipping box}
  \end{minipage}\hfill  \begin{minipage}[t]{0.494\linewidth}\centering
    \includegraphics[width=0.494\linewidth]{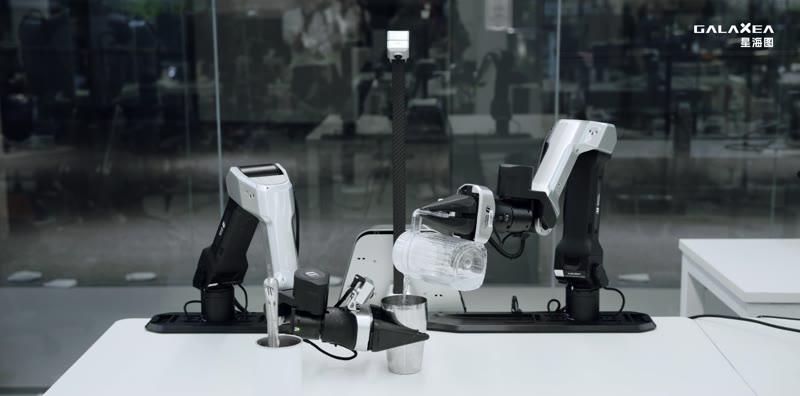}\hfill
    \includegraphics[width=0.494\linewidth]{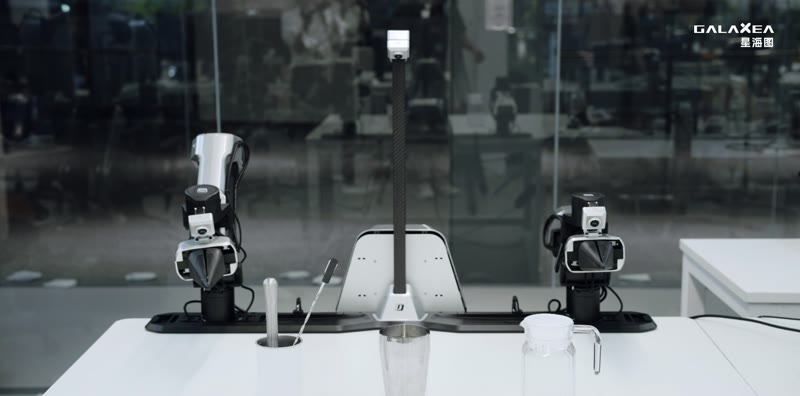}\\[1pt]
    {\footnotesize Make lemon tea}
  \end{minipage}\par\vspace{5pt}
    \begin{minipage}[t]{0.494\linewidth}\centering
    \includegraphics[width=0.494\linewidth]{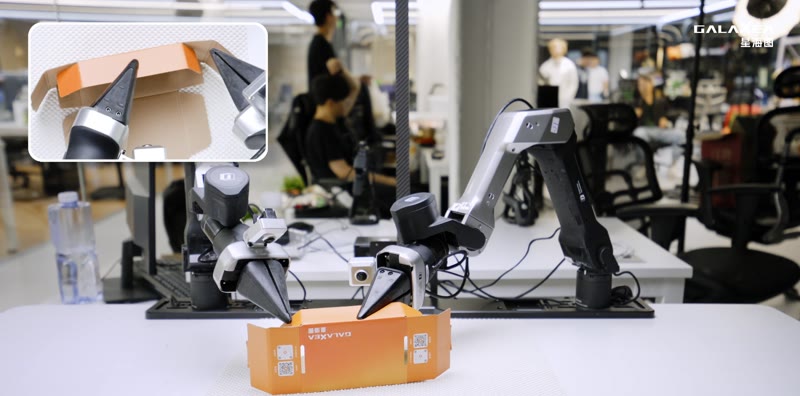}\hfill
    \includegraphics[width=0.494\linewidth]{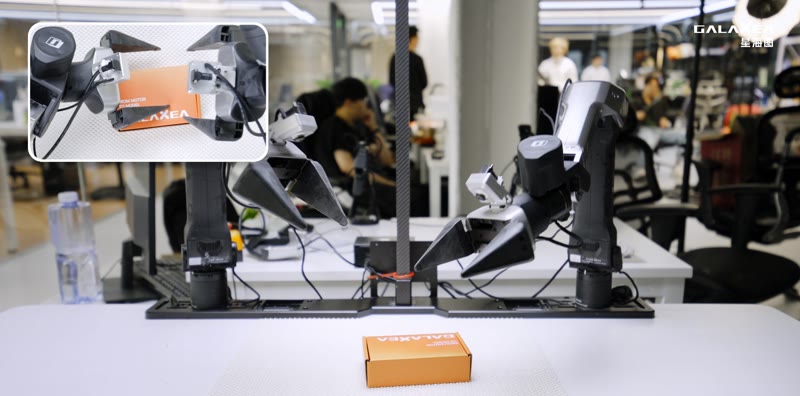}\\[1pt]
    {\footnotesize Assemble a gift box}
  \end{minipage}\hfill  \begin{minipage}[t]{0.494\linewidth}\centering
    \includegraphics[width=0.494\linewidth]{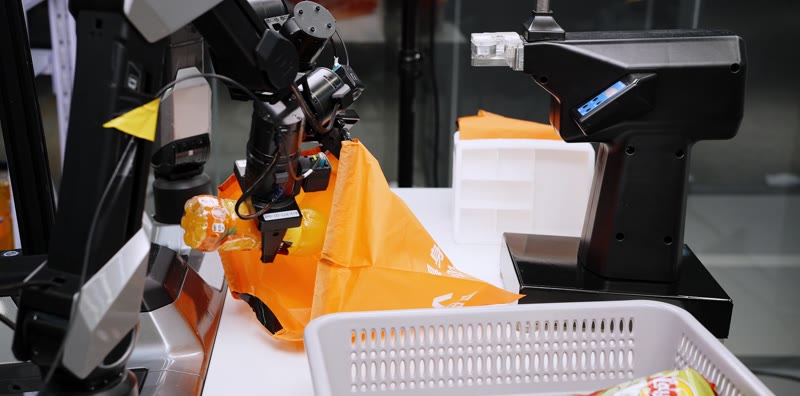}\hfill
    \includegraphics[width=0.494\linewidth]{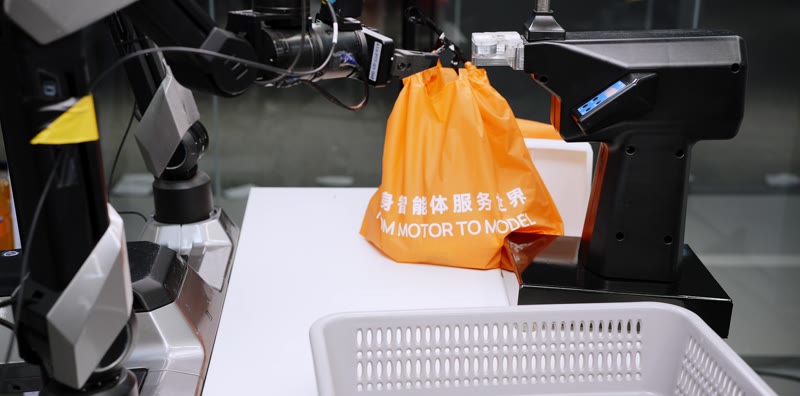}\\[1pt]
    {\footnotesize Bag products}
  \end{minipage}\par\vspace{5pt}
    \begin{minipage}[t]{0.244\linewidth}\centering
    \includegraphics[width=\linewidth]{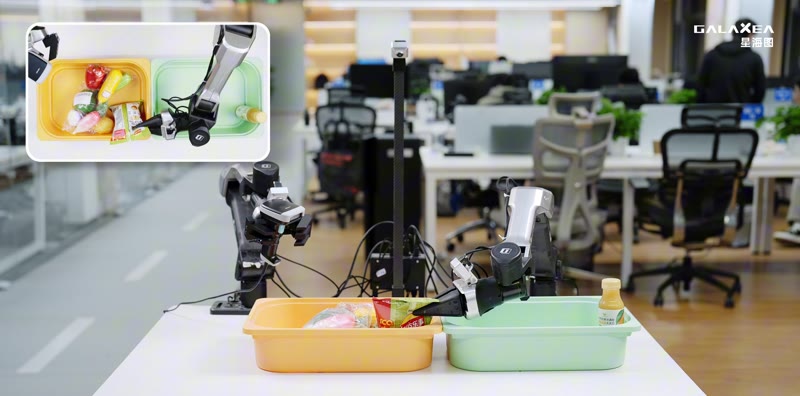}\\[1pt]
    {\footnotesize Scan a product}
  \end{minipage}\hfill  \begin{minipage}[t]{0.244\linewidth}\centering
    \includegraphics[width=\linewidth]{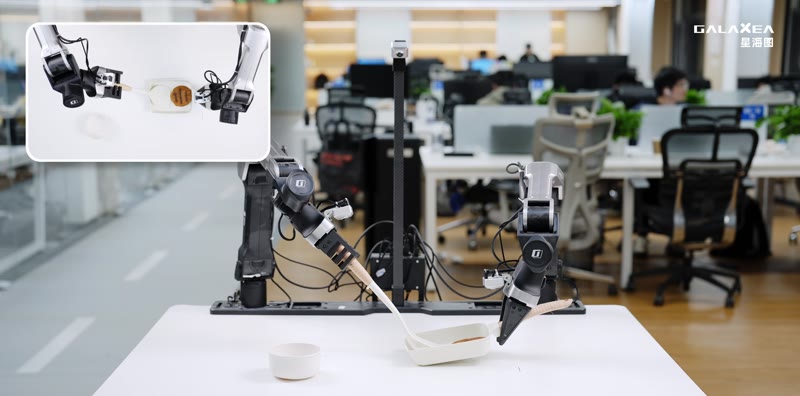}\\[1pt]
    {\footnotesize Scoop with a spatula}
  \end{minipage}\hfill  \begin{minipage}[t]{0.244\linewidth}\centering
    \includegraphics[width=\linewidth]{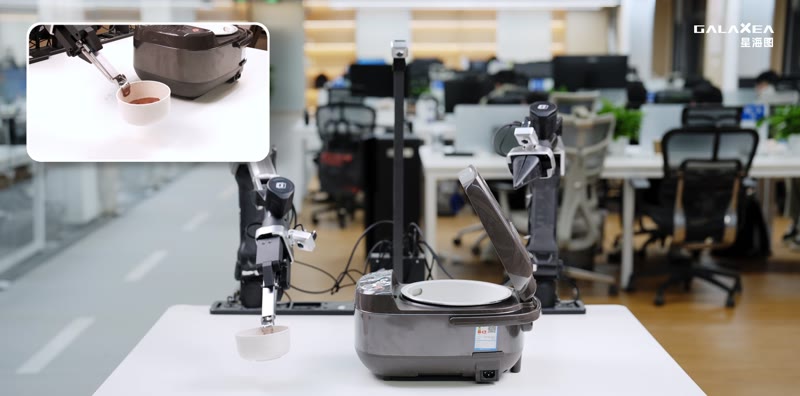}\\[1pt]
    {\footnotesize Lift a bowl with tongs}
  \end{minipage}\hfill  \begin{minipage}[t]{0.244\linewidth}\centering
    \includegraphics[width=\linewidth]{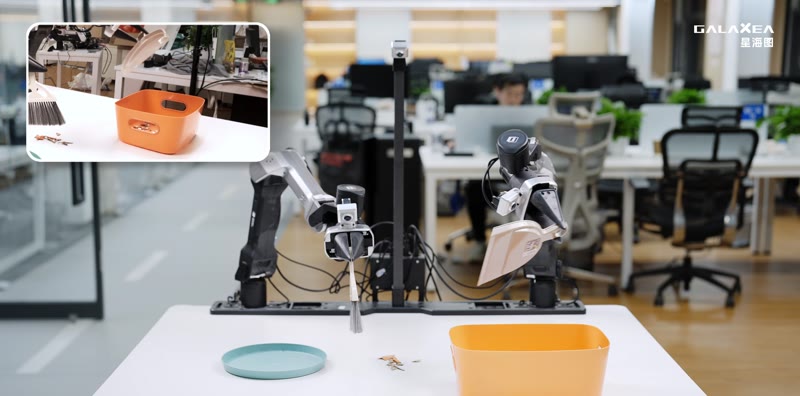}\\[1pt]
    {\footnotesize Sweep the table}
  \end{minipage}
  \caption{Post-training tasks. Top two rows: long-horizon production tasks, each shown mid-execution and at completion, learned from roughly five hours of post-training data per task. Bottom row: the four tool-use tasks, learned from 20 demonstrations per tool. Frames are from autonomous rollouts.}
  \label{fig:posttrain-rollouts}
\end{figure}

Finally, we evaluate how efficiently \ours{} adapts to new physical skills across two regimes: few-shot \emph{tool use} and \emph{long-horizon production tasks} (Figure~\ref{fig:posttrain-rollouts}). For tool use, we collect 20 demonstrations per tool across four tasks: scooping food with a spatula, lifting a bowl with tongs, scanning a barcode with a handheld scanner, and sweeping a table with a small broom. For long-horizon production tasks, we post-train on roughly five hours of data per task to fold shipping and gift boxes from flat cardboard, make lemon tea, and bag retail products. These workflows run for several minutes and involve deformable materials and liquids, where compounding errors easily cause failure.

All models are post-trained on identical datasets and evaluated by both overall success rate and task progress (the fraction of task stages completed per trial).

\begin{figure}[!htb]
  \centering
  \includegraphics[width=\linewidth]{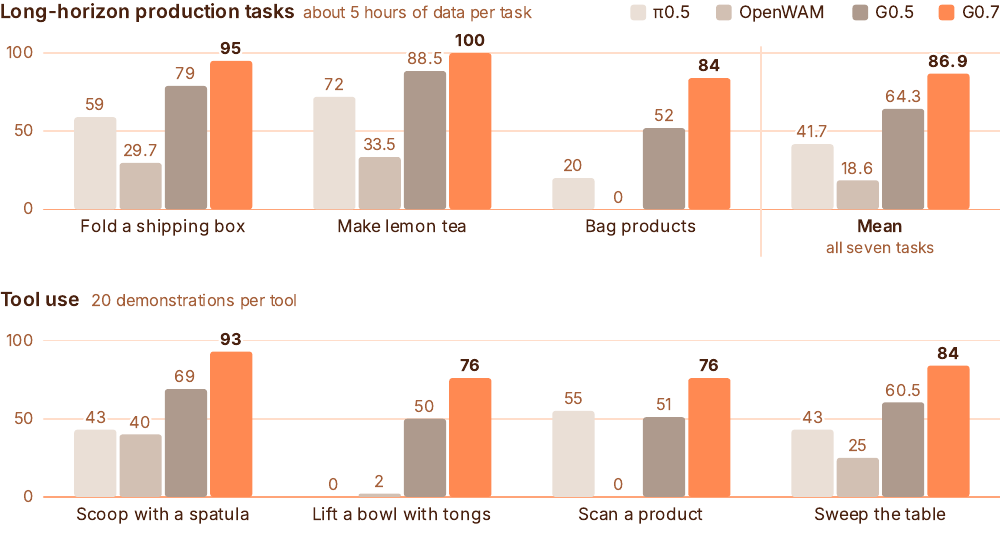}
  \caption{Task progress (\%) after real-robot post-training. Top: three long-horizon production tasks (roughly five hours of data each) and the mean across all seven evaluated tasks. Bottom: four tool-use tasks (20 demonstrations per tool).}
  \label{fig:posttrain}
\end{figure}

\paragraph{Long-horizon production tasks.}
Across all seven evaluated post-training tasks, \ours{} achieves 86.9\% mean task progress, compared with 64.3\% for G0.5, 41.7\% for \pifive{}, and 18.6\% for OpenWAM (Figure~\ref{fig:posttrain}). On shipping-box folding, \ours{} achieves a 90\% success rate and 95\% task progress from five hours of demonstrations, compared with 75\% success (79\% progress) for G0.5 and 55\% success (59\% progress) for \pifive{}, while OpenWAM fails to complete a full box (29.7\% progress). On lemon tea, \ours{} completes every trial (100\% success and task progress), compared with 80\% success (88.5\% progress) for G0.5 and 45\% success (72\% progress) for \pifive{}, while OpenWAM never finishes the drink (33.5\% progress). On product bagging, \ours{} reaches 63\% success and 84\% task progress, compared with 50\% success (52\% progress) for G0.5 and 20\% success and progress for \pifive{}, while OpenWAM makes no progress.

\paragraph{Few-shot tool use.}
From 20 demonstrations per tool, \ours{} reaches 82\% mean task progress across the four tools, outperforming G0.5 (58\%), \pifive{} (35\%), and OpenWAM (17\%). \ours{} maintains high progress across all four tools (93\% on spatula scooping, 84\% on table sweeping, and 76\% on both tong grasping and barcode scanning), whereas \pifive{} and OpenWAM make little or no progress (0\%--2\%) on precision-critical tasks such as lifting a bowl with tongs.

\subsection{Data Scaling}
\label{sec:scaling}

\begin{figure}[!htb]
  \centering
  \includegraphics[width=\linewidth]{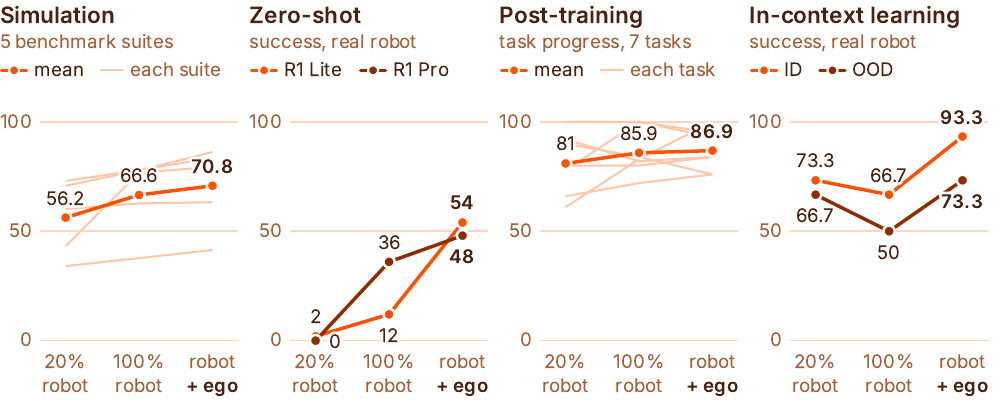}
  \caption{Scaling performance across the three pretraining checkpoints (\oursk{}, \oursr{}, and \ours{}) in simulation, real-robot zero-shot deployment, real-robot post-training, and in-context learning.}
  \label{fig:scaling}
\end{figure}

Performance improves across all four evaluation suites as pretraining scales from \oursk{} to \oursr{} and \ours{} (Figure~\ref{fig:scaling}). Expanding the robot corpus from \oursk{} to \oursr{} lifts the five-suite simulation mean from 56.2 to 66.6 and real-robot post-training progress from 81.0\% to 85.9\%; adding ego video in \ours{} brings them to 70.8 and 86.9\%, respectively. Zero-shot deployment depends on scale the most: \oursk{} almost never succeeds without task data (0 of 25 trials on R1 Pro and 1 of 50 on R1 Lite). Ego data has the largest impact where the policy must generalize beyond its training distribution: from \oursr{} to \ours{}, zero-shot success on R1 Lite jumps from 12\% to 54\%, and out-of-distribution success on in-context learning tasks rises from 50\% to 73\%. Taken together, these trends reveal complementary roles for the two data sources: scaling robot teleoperation steadily improves in-distribution proficiency and downstream post-training performance, while adding human egocentric video unlocks out-of-distribution generalization across zero-shot deployment and video-conditioned in-context learning that robot data alone does not reach.

\section{Related Work}
\label{sec:related}

\paragraph{Vision-language-action and world action models.}
Vision-language-action models (VLAs) such as $\pi_0$~\citep{pi0} and \pifive{}~\citep{pi05} adapt pretrained vision-language backbones to output continuous robot actions via flow matching or autoregressive tokenization. World action models (WAMs) instead initialize from pretrained video generation models and co-train future video prediction alongside action decoding to transfer physical dynamics priors into control~\citep{dreamzero,lingbot,lingbot2,vpp2,openwam,fastwam}. Most recent WAMs, including DreamZero~\citep{dreamzero}, LingBot-VA~\citep{lingbot}, and VPP2~\citep{vpp2}, follow an imagine-then-act paradigm at deployment, denoising future video frames before predicting the corresponding actions. Fast-WAM~\citep{fastwam} demonstrated that video supervision is crucial during representation learning but unnecessary at test time. \ours{} scales this finding to long-context autoregressive pretraining on large-scale data while retaining both direct action decoding (Fast mode) and imagine-then-act inference (IDM mode) within a single checkpoint.

\paragraph{Long-context, memory, and in-context robot policies.}
Standard visuomotor policies condition on only one or two recent observations, making them memoryless under partial observability. Recent WAMs have explored extending temporal context through persistent key-value (KV) caching across rollout chunks~\citep{lingbot,lingbot2}, studying autoregressive context scaling~\citep{longwam}, or introducing dedicated memory modules that compress history into anchor and gist tokens~\citep{memorywam}. \ours{} adopts a streamlined architectural path: it maintains a sliding-window KV cache containing only clean latents from real camera observations, trains autoregressively over multi-chunk sequences from the beginning of pretraining without auxiliary memory heads, and extends its context window during downstream ICL adaptation to follow one-shot human video demonstrations and track hidden state over multi-minute horizons.

\paragraph{Low-latency action chunking and asynchronous control.}
Predicting multi-step action chunks improves temporal consistency in high-frequency manipulation, but synchronous control loops still pause at chunk boundaries while waiting for model inference, a bottleneck that is exacerbated when WAMs must first denoise future video latents. Real-time chunking (RTC)~\citep{rtc} eliminates inter-chunk pauses for flow-matching policies at inference time by denoising the next chunk asynchronously while the current chunk executes, freezing committed prefix steps and inpainting the remaining horizon. Because \ours{}'s Fast mode bypasses video generation and reduces inference to a 60\,ms flow-matching action denoising pass over cached context latents, it serves as a drop-in fit for RTC without modifying the training objective.

\paragraph{Learning manipulation from human egocentric video.}
Collecting first-person human manipulation video is significantly more scalable than robot teleoperation and captures much broader scene, object, and behavioral diversity. Prior work has leveraged human video by learning latent action spaces from action-free video~\citep{lapa,moto}, or by estimating 3D hand and wrist trajectories and mapping them into robot action spaces~\citep{egomimic,egovla,beingh0,egodex}. \ours{} follows the explicit 3D kinematic alignment route at scale: we extract 6-DoF wrist poses aligned to the tool-center point (TCP) axis convention of our robot grippers together with 3D five-fingertip positions, treating human hands as an additional embodiment co-trained alongside robot teleoperation under the same joint video-action objective.

\section{Conclusion}
\label{sec:conclusion}

\ours{} extends the Fast-WAM recipe to context windows of up to five minutes and demonstrates that direct action decoding remains effective after large-scale pretraining on robot and human ego data. With video prediction as a training objective, Fast mode performs comparably to IDM on the same checkpoint and reduces inference latency to 60\,ms, enabling smooth closed-loop execution with real-time chunking. Native long-context pretraining equips the policy with both episode memory and video-based in-context learning, while the combined robot and ego corpus provides a strong foundation for zero-shot deployment and rapid adaptation to complex real-world manipulation tasks.

\section*{Contributions}
\paragraph{Model design and pretraining.} Tianyuan Yuan, Baijun Ye, Zijie Zhao, Zibin Dong, Haonan Liu, Feiyang Xue, Yicheng Liu.
\paragraph{Post-training and experiments.} Shiduo Zhang, Xiaopeng Yu, Yizhao Xu, Baijun Ye, Zijie Zhao, Zheng Liu, Haobo Jia, Bowen Zhang, Haodong Yang, Haoyu Zhang, Yang Wen.
\paragraph{Data.} Xing Nie, Shicheng Cao, Ke Dong, Anqi Yang, Jianpeng Yu, Shenling Qiu, Haobo Jia, Tailai Cheng, Longhao Wei.
\paragraph{Policy infrastructure.} Tianyuan Yuan, Anqi Yang, Jiahui Niu, Yue Sun.
\paragraph{Project lead.} Tianyuan Yuan.
\paragraph{Project PI.} Hang Zhao.

\bibliographystyle{plainnat}
\bibliography{references}

\end{document}